\documentclass[conference]{IEEEtran}
\IEEEoverridecommandlockouts
\usepackage{cite}
\usepackage{amsmath,amssymb,amsfonts}
\usepackage{algorithmic}
\usepackage{graphicx}
\usepackage{textcomp}
 \usepackage{multirow}
 \usepackage[table]{xcolor}
\usepackage{makecell}
\def\BibTeX{{\rm B\kern-.05em{\sc i\kern-.025em b}\kern-.08em
    T\kern-.1667em\lower.7ex\hbox{E}\kern-.125emX}}
\begin{document}

\title{RAL:
Redundancy-Aware Lipreading Model Based on Differential Learning with Symmetric Views\\
}

\author{\IEEEauthorblockN{1\textsuperscript{st} Zejun Gu}
\IEEEauthorblockA{
\textit{Hefei University of Technology}\\
Hefei, China \\
guzejunmail@gmail.com}
\and
\IEEEauthorblockN{2\textsuperscript{nd} Junxia Jiang}
\IEEEauthorblockA{
\textit{Hefei University of Technology}\\
Hefei, China \\
2021111089@mail.hfut.edu.cn
}
}


\maketitle

\begin{abstract}
Lip reading involves interpreting a speaker's speech by analyzing sequences of lip movements. Currently, most models regard the left and right halves of the lips as a symmetrical whole, lacking a thorough investigation of their differences. However, the left and right halves of the lips are not always symmetrical, and the subtle differences between them contain rich semantic information. In this paper, we propose a differential learning strategy with symmetric views (DLSV) to address this issue. 
Additionally, input images often contain a lot of redundant information unrelated to recognition results, which can degrade the model's performance. We present a redundancy-aware operation (RAO) to reduce it. Finally, to leverage the relational information between symmetric views and within each view, we further design an adaptive cross-view interaction module (ACVI). Experiments on LRW and LRW-1000 datasets fully demonstrate the effectiveness of our approach.
\end{abstract}


\section{Introduction}
Lip reading aims to predict spoken content solely by analyzing videos of speaking faces. Lip reading has a broad array of applications, including generating lip patterns for security authentication~\cite{7310809}, improving the reliability of human-computer interactions~\cite{10.1007/978-3-319-25903-1_49}, creating highly realistic talking avatars~\cite{10.1145/3072959.3073640}, and aiding communication for individuals with hearing impairments~\cite{duchnowski1994see}. 

\begin{figure}[t]
\begin{center}
\setlength{\belowcaptionskip}{-0.7cm}
   \includegraphics[width=1.0\linewidth]{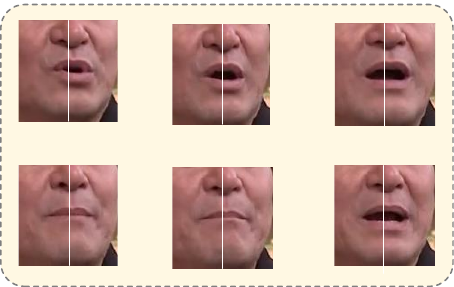}
\vspace{-2.0em}
\caption{Comparison between left and right half-lip views in lip-reading video frames. In many lip-reading images, there are significant differences between the left and right half-lip views.
}
\label{mouth}%
\end{center}
\vspace{-2.0em}
\end{figure}

Currently, many lip reading studies focus on utilizing spatial information\cite{martinez2020lipreading,tian2022lipreading}, such as Temporal Convolutional Networks\cite{martinez2020lipreading}, whole-part collaborative learning~\cite{tian2022lipreading}.
However, these methods lack attention to the symmetrical left and right half-lip views.
In real-world scenarios, the left and right half-lips in a video sequence are not always the same. as shown in Figure~\ref{mouth}.
For example, when a person experiences emotional fluctuations, the left and right half-lips may exhibit different changes.
These differences, as a form of micro-expression, carry rich semantic information. 
Additionally, in continuous speech videos, these differences typically occur during word transitions, which contain important inter-frame associative information.

Furthermore, most current models ignore the redundant information present in the input frames, which is often background information unrelated to the lips. This not only fails to aid lip reading but also consumes computational resources. It significantly affects the model's learning performance.

To address the first challenge, we propose a differential learning strategy with symmetric views (DLSV). Specifically, we first divide the image features into left-view and right-view features. Then, the left and right view features are separately fed into the DRSBlock, 
which has shared weights, to learn their identical information. Finally, an adaptive cross-view interaction module (ACVI) is used to effectively capture the relationships between different areas of the lips and extract their differential
information. Through repeated feature extraction, not only are the identical features of the left and right views enhanced, but their differences are also fully learned.

To address the second challenge, we present a redundancy-aware operation (RAO). Inspired by DRSN\cite{zhao2019deep}, we employ a sub-network that integrates an attention mechanism and soft threshold function to construct the redundancy-aware operation. The attention mechanism can determine the threshold vector of the threshold function based on the degree of redundancy in the feature content. By using the soft threshold function, the redundant information can be filtered more effectively.
\begin{figure}
\begin{center}
\setlength{\belowcaptionskip}{-0.5cm}
\includegraphics[width=1.00\linewidth,height=0.58\linewidth]{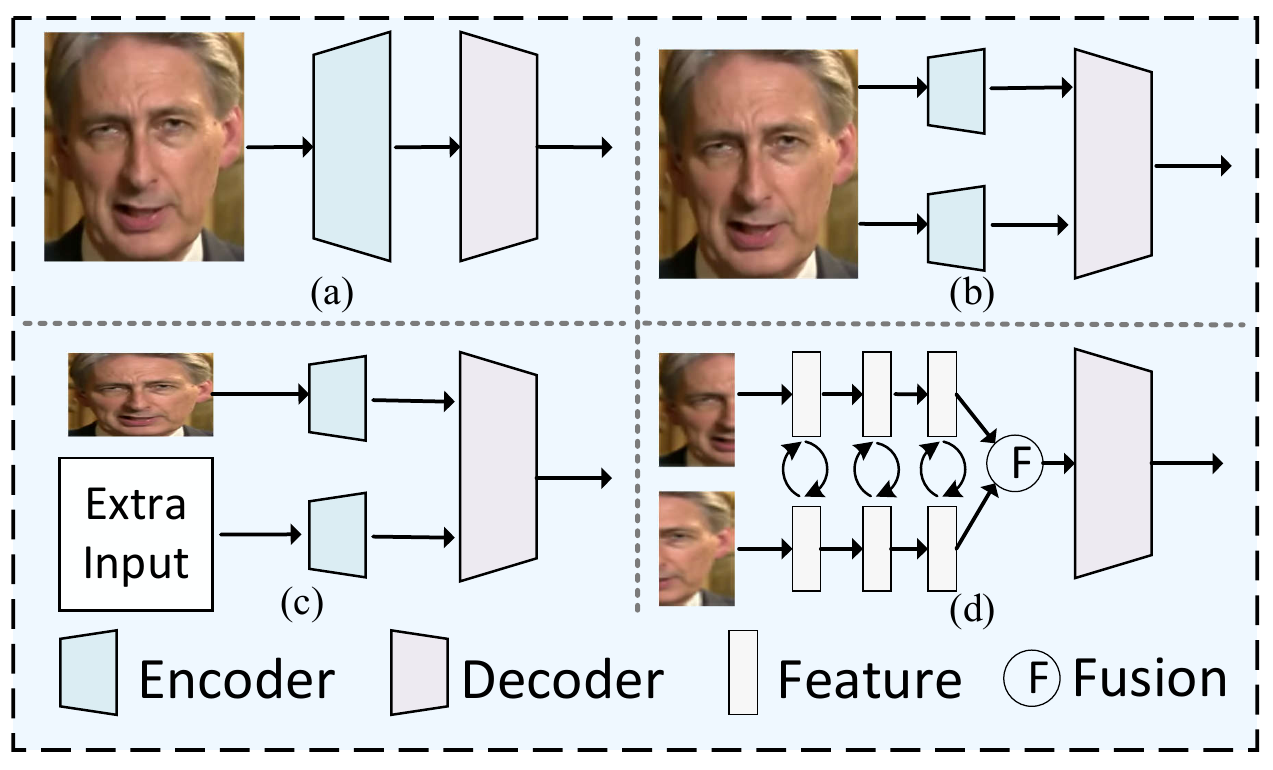} 
\vspace{-2.0em}
\caption{Comparison of Different Lip Reading Model Architectures. (a) TCN. (b) WPCL. (c) UDP. (d) our RAL model. }
\label{arch}
\end{center}
\vspace{-2.0em}
\end{figure}
In lip-reading recognition, the positional relationships between different parts of the lips are crucial. We design an adaptive cross-view interaction module to make full use of the relational information of each part, thereby enabling further learning of differential information.

The contributions of our work are:
\begin{itemize}
\item  We propose a novel lip-reading model, RAL, which is the first work to focus on the identical and different information of the symmetrical left and right halves of the lips.
\item  
We introduce a redundancy-aware operation that dynamically adjusts the threshold vector based on the input content to filter out redundant information.
\item  We design an adaptive cross-view interaction module to learn both visual clues and constraint relations in symmetric views.
\item We show the effectiveness of the proposed model by applying it to various language datasets.
Considerable improvements are observed in these
datasets.
\end{itemize}
\section{Related Work}
With advancements in deep learning and speech processing technologies, lip reading has achieved significant progress. 
Such progress could be made with improved neural network architecture\cite{petridis2018end}, large-scale audio-visual datasets\cite{zhao2019cascade},  enhanced multi-modal learning strategies\cite{Kim2022DistinguishingHU}, and carefully crafted training methods\cite{Ma2022VisualSR}. 

\cite{stafylakis2017combining} introduces an architecture combining ResNet\cite{he2016deep} and RNN\cite{graves2012long} to enhance word-level lip reading performance. 
Some approaches tend to fully utilize temporal information, such as optical flow\cite{Ma2022VisualSR}, dynamic flow\cite{Martnez2020LipreadingUT}, and time-shift modules\cite{Assael2016LipNetES}. \cite{Ma2022VisualSR} proposes to incorporate optical flow information alongside RGB information, using a two-stream network to encode them. \cite{Martnez2020LipreadingUT} replaces the RNN-based back-end architecture with temporal convolutions, significantly improving word-level lip reading performance. 
In addition to word-level lip reading, \cite{Assael2016LipNetES} presents an end-to-end sentence-level lip reading model that leverages Connectionist Temporal Classification (CTC)\cite{GravesConnectionistTC}. 
\cite{Almajai2016ImprovedSI} addresses the issue of speaker dependency and proposes speaker-adaptive lip reading models.
\cite{Afouras2019ASRIA} utilizes knowledge distillation to transfer knowledge from a superior model to a student model. \cite{Kim2022CroMMVSRCM} designs using memory networks to incorporate auditory knowledge into lip reading models without requiring audio inputs.  Furthermore, some self-supervised training methods for pre-training neural networks have demonstrated outstanding performance in lip reading \cite{Shi2022LearningAS, Chung2016OutOT}. 
\section{Method}

\subsection{Overview}
The overall structure of our redundancy-aware lipreading (RAL) model is depicted in Figure~\ref{zong}. It consists of a backbone (3DCNN)\cite{Liao20183DCN}, a feature encoder, several adaptive cross-view interaction (ACVI)  modules, and a temporal decoder based on
MSTCN\cite{kim2022speaker} for temporal modeling. 

\subsection{Differential learning strategy with symmetric views}
\begin{figure*}
\begin{center}
\setlength{\belowcaptionskip}{-0.5cm}
\includegraphics[width=0.99\linewidth,height=0.44\linewidth]{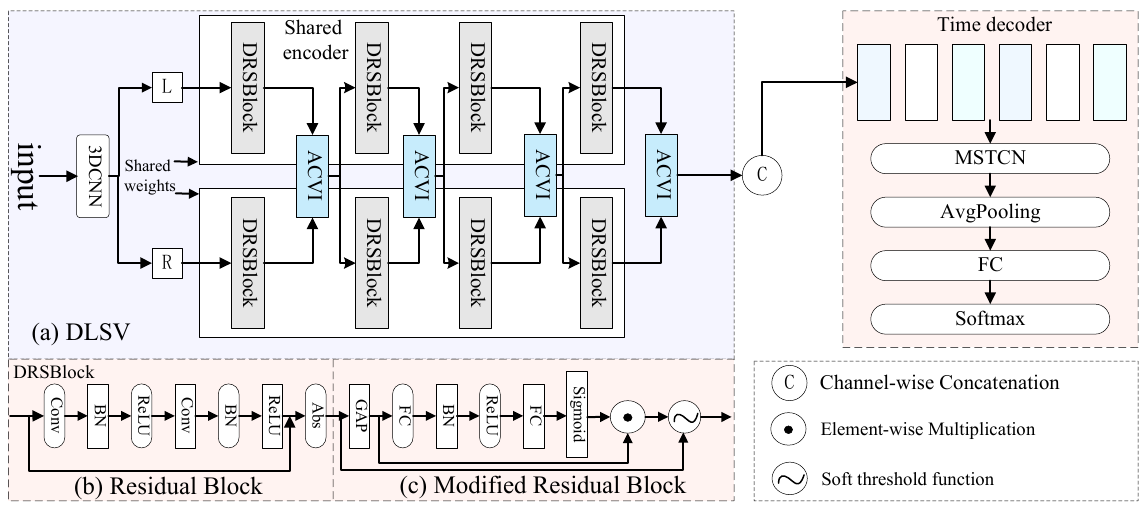}  
\vspace{-0.8em}
\caption{Schematic illustration of the proposed RAL model.
}
\label{zong}
\end{center}
\vspace{-2.0em}
\end{figure*}

As shown in Figure~\ref{arch}, previous works have tried to introduce multi-branch approaches to fully leverage the spatial information in lip-reading images. such as TCN\cite{martinez2020lipreading}, WPCL ~\cite{tian2022lipreading}, etc.
However, they do not fully focus on the identical and differential information in the left-right half-lip,
which contains rich semantic information. Hence, we propose a differential learning strategy with symmetric views (DLSV) to solve this problem (shown in Figure~\ref{zong}).

Specifically, we first use a 3DCNN to extract the initial features, and then segment the lip features into left and right half-lip features. A shared encoder with shared weights is employed to extract the common features of the left and right half-lips, and an adaptive cross-view interaction module is used to further capture the relationships between them. The ACVI module further enhances the unique feature of the left and right half-lips. Through multiple iterations of segmentation, extraction, and interaction, the differential information is fully learned.
As shown in Figure~\ref{zong}(a), the shared encoder consists of several DRSBlocks. Each DRSBlock is composed of a residual block and a modified residual block based on redundancy-aware operations.
\subsection{Redundancy-aware operation}
To remove redundant information unrelated to lip-reading recognition, we design a redundancy-aware operation. 
Inspired by DRSN\cite{zhao2019deep}, the redundancy-aware operation is implemented through a sub-network that utilizes an attention mechanism and a soft threshold function. 
The attention mechanism is designed to adaptively generate the threshold vector of the soft threshold function based on the image content.
The detailed design of RAO is illustrated in Figure~\ref{zong}(c).

Firstly, the feature map $X\in \mathbb{R}^{ C\times H\times W} $ is converted into a one-dimensional vector using an absolute value operation ($Abs$) followed by a global average pooling layer ($GAP$), and is then processed through a two-layer fully connected ($FC$) network.

 \begin{equation}
     F=FC(GAP\left (  Abs(X) \right ) ),
 \end{equation}
where $F\in \mathbb{R}^{ C\times 1\times 1} $ 
represents the feature extracted from the two-layer fully connected ($FC$) network. To ensure the outputs of the $FC$ network are within an appropriate range, they are scaled using a sigmoid function \cite{lecun2015deep}.

\begin{equation}
    \sigma =\frac{1}{1+e^{-F} } ,
\end{equation}
 where $\sigma $ denotes the scaling parameters.
Subsequently, the threshold is calculated as follows:
\begin{equation} 
    \tau =\sigma \odot GAP(Abs(X)),
\end{equation}
where $\tau\in \mathbb{R}^{ C\times 1\times 1}$ refers the threshold, and $\odot$ refers an element-wise multiplication operation.
It is worth noting that the threshold we obtain is not a single value but rather a vector. 
The operation ensures that the threshold remains positive and within a reasonable range, preventing them from becoming excessively large and resulting in all-zero feature outputs. 

Finally, the original feature map $X$ is combined with the threshold from the attention mechanism to perform the soft threshold operation.



\subsection{Adaptive cross-view interaction module}
In lip-reading recognition, the relational information of various parts of the lips is very important. For example, the positional relationships of different parts, and the similarities and differences in symmetrical positions. 
We propose an adaptive cross-view interaction module to fully utilize the relational information between symmetric views and within each view.

\begin{figure}[t]
\begin{center}
\setlength{\belowcaptionskip}{-8cm}

\includegraphics[width=0.99\linewidth,height=1.08\linewidth]{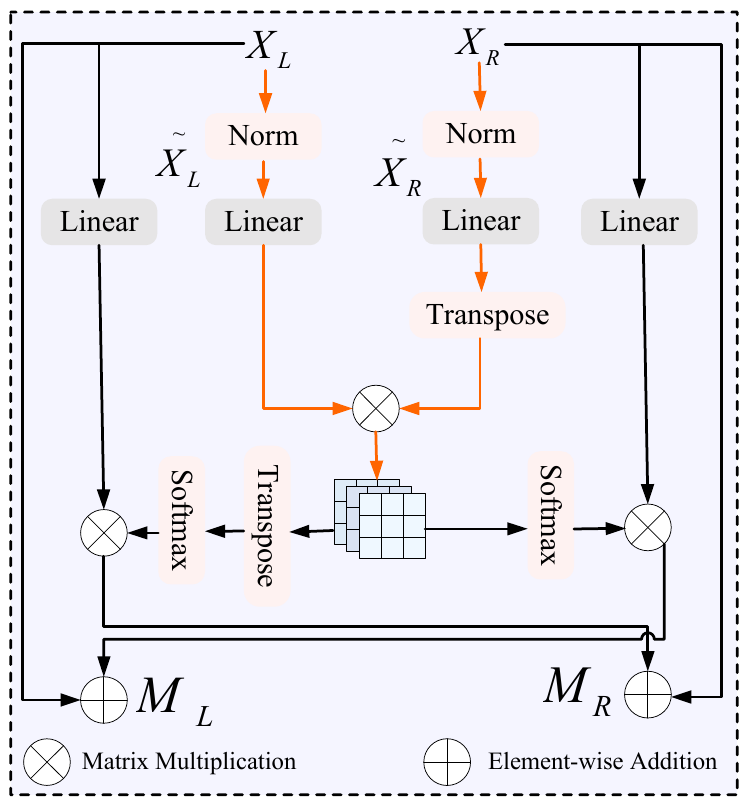}
\vspace{-2.0em}
\caption{The structure of our proposed ACVI module.
}
\label{acvi}
\end{center}
\vspace{-1.0em}
\end{figure}

The structure of the proposed ACVI is shown in Figure~\ref{acvi}.
It uses scaled dot-product attention to compute the dot products of the query with all keys and applies a softmax function to obtain the weights for the values.:
\begin{equation}
    Attention\left ( Q,K,V \right ) =softmax\left ( QK^{T}/\sqrt{C}    \right ) V,
\end{equation}
where $ Q\in \mathbb{R}^{H\times W \times C}$ is the query matrix projected from the left-view feature (or right-view feature), and $ K, V\in \mathbb{R}^{H\times W \times C}$ are key, value matrices projected by
the other view (right-view or left-view). 
In detail, given the input features $X_{L},X_{R} \in \mathbb{R}^{H\times W \times C} $, we get layer normalized features $\tilde{X}_{L}  =LN\left ( X_{L}  \right ) $ and $\tilde{X}_{R}  =LN\left ( X_{R}  \right ) $. Then, we calculate bidirectional cross attention between left and right views, to  facilitate the interaction between them, by:
\begin{equation}
    M_{R\rightarrow L }=Attention\left ( W_{1}^{L}\tilde{X}_{L} , W_{1}^{R}\tilde{X}_{R},W_{2}^{R}X_{R} \right ) ,
\end{equation}
\begin{equation}
    M_{L\rightarrow R }=Attention\left ( W_{1}^{R}\tilde{X}_{R} , W_{1}^{L}\tilde{X}_{L},W_{2}^{L}X_{L} \right ) ,
\end{equation}
where $W_{1}^{L}$, $W_{1}^{R}$, $W_{2}^{L}$ and $W_{2}^{R}$ are corresbonding projection matrices. Finally, the interaction
information feature
$M_{R\rightarrow L }$, $M_{L\rightarrow R }$ and intra-view shared information ${X}_{L}$, ${X}_{R}$ are fused by element-wise addidion:
\begin{equation}
    M_{L}= \alpha _{L} M_{R\rightarrow L }+X_{L} ,
\end{equation}
\begin{equation}
    M_{R}= \alpha _{R} M_{L\rightarrow R }+X_{R} ,
\end{equation}
where $M_{L}$ and $M_{R}$ are output features from the left view and the right view, $\alpha _{L}$ and $\alpha _{R}$ 
are trainable 
scale factors. 


\section{Experiments}
\subsection{Experimental Settings}
\subsubsection{Datasets}
The LRW dataset\cite{Chung2016LipRI} consists of 500 English word classes, each containing 1,000 samples. 
It includes some similar words, such as those with singular and plural forms or tense variations. 
The LRW-1000 dataset\cite{yang2019lrw}~includes more than 1,000 Chinese word classes, each consisting of one or more Chinese words, totaling over 70,000 samples. 

\subsubsection{Preprocessing}

For both the LRW and LRW-1000 datasets, each sample is cropped to a fixed Region of Interest (RoI) of 96×96 pixels. All images are converted to grayscale to reduce computational complexity. During training, each frame is randomly cropped to 88×88 pixels.

\subsubsection{Implementations}
\indent 
All models are implemented on Pytorch 
and trained on two NVIDIA RTX 2080ti GPUs with 11GB
memory. We use a cosine scheduler and Adam optimizer. 
The initial learning rate is 3e-4 and the weight decay is 1e-4.
In addition, to enhance the robustness of the temporal model, we use the variable length enhancement strategy \cite{martinez2020lipreading}.
\subsection{Experimental Results}
\begin{table}[t]
\begin{center}
\caption{Comparison with other related works on
LRW dataset. The proposed model achieves
competitive performance to other methods.}
\vspace{-2.0em}
\setlength{\tabcolsep}{7.0pt}
\renewcommand{\arraystretch}{1.2}
\resizebox{\linewidth}{!}{
\setlength{\tabcolsep}{9mm}{\begin{tabular}{cc}
\Xhline{1.2pt}
\textbf{Method} &  \textbf{ACC($\%$)} \\
\hline
Multi-Grained \cite{wang2019multi} & 83.3  \\
BiGRU+GRU\cite{luo2020pseudo} & 83.5 \\
Deformation Flow \cite{xiao2020deformation} & 84.1  \\
BiGRU+Cutout \cite{zhang2020can} & 85.0 \\
TSM \cite{hao2021use} & 86.2   \\
HPNet+Self-Attention\cite{chen2021automatic}&86.8\\
LiRA\cite{ma2021lira} & 88.1 \\
 DC-TCN\cite{ma2021lip} & 88.4   \\
MSTCN-Ensemble \cite{ma2021towards} & 88.5   \\
MVFN\cite{zhang2022boosting}&87.4\\
UDP\cite{kim2022speaker}&87.9\\
WPCL~\cite{tian2022lipreading} &88.3\\
UDP+SI\cite{kim2022speaker}&88.5\\
\rowcolor{gray!10}MS-TCN (baseline)\cite{martinez2020lipreading} & 85.3   \\
Ours & \textbf{89.3}\textcolor{blue}{$_{
    +4.0}$}  \\
\Xhline{1.2pt}
\end{tabular}}
}
\vspace{-2.0em}
\label{per01}
\end{center}
\end{table}

\begin{table}[t]
\vspace{-1.0em}
\begin{center}
\caption{Comparison with other related works on
LRW-1000 dataset. 
The proposed model achieves
significant performance to other methods.
}
\vspace{-1.0em}
\setlength{\tabcolsep}{11.0pt}
\renewcommand{\arraystretch}{1.4}
\resizebox{\linewidth}{!}{
\setlength{\tabcolsep}{10mm}{\begin{tabular}{cc}
\Xhline{1.2pt}
\textbf{Method} &  \textbf{ACC($\%$)} \\
\hline
Multi-Grained \cite{wang2019multi}  & 36.9  \\
BiGRU+GRU\cite{luo2020pseudo} & 38.7 \\
Deformation Flow \cite{xiao2020deformation}  & 41.9  \\
TSM \cite{hao2021use}  & 44.6  \\
DC-TCN\cite{ma2021lip}  & 43.7  \\
\rowcolor{gray!10}MS-TCN (baseline)\cite{martinez2020lipreading} & 41.4  \\
Ours & \textbf{46.5}\textcolor{blue}{$_{
    +5.1}$}  \\
\Xhline{1.2pt}
\end{tabular}}
}
\label{lrw1000}
\end{center}
\vspace{-2.0em}
\end{table}

\begin{table}[htbp]
  \centering
  \setlength{\tabcolsep}{9.0pt}
  \renewcommand{\arraystretch}{1.1}
  \caption{Ablation study for 
each proposed component on the LRW dataset.}
\vspace{-1.0em}
\resizebox{\linewidth}{!}{
    \begin{tabular}{ccccc}
    \Xhline{1.2pt}
    Method & DLSV  & RAO   & ACVI  & ACC(\%) \\
    \hline
    \rowcolor{gray!10}Baseline &  -     &    -   &   -    & 85.3 \\
    Ours  & \checkmark     &   -    &   -    & 86.8\textcolor{blue}{$_{+1.5}$}\\
    Ours  & \checkmark     & \checkmark     & -      & 87.8\textcolor{blue}{$_{
    +2.5}$} \\
    Ours  &    -   &   -    & \checkmark     & 88.3\textcolor{blue}{$_{
    +3.0}$} \\
    Ours  & \checkmark     & \checkmark     & \checkmark     & \textbf{89.3}\textcolor{blue}{$_{
    +4.0}$} \\
    \Xhline{1.2pt}
    \end{tabular}%
    }
    \vspace{-0.5em}
  \label{ablation}%
\end{table}%

\subsubsection{Comparison with other related works}


To study the performance of the proposed RAL model, we
compare it with other lip-reading models.
The experimental results on the LRW dataset are shown in Table~\ref{per01}.
As shown, the proposed model brings significant improvement (4.0\%) to the baseline method. 
Results presented in Table~\ref{per01} demonstrate that
our method shows consistent performance superiority over related works, which provides strong evidence of the effectiveness of our method.
We conduct comparisons on LRE-1000
dataset and present the results in Table~\ref{lrw1000}.  
Our proposed RAL model surpasses the TSM and DC-TCN by 1.9\% and 2.8\%, respectively. It achieves the best performance on both the LRW dataset and the LRW-1000 dataset.

\subsubsection{Ablation Study}
\indent 
In this subsection, we conduct several ablation experiments
to show how each proposed component helps the model improve its performance.
As shown in Table~\ref{ablation}, all the proposed components benefit the
lip-reading model. 
DLSV and ACVI each bring an
improvement of 1.5\% and 3.0\%, respectively. 
The differential learning
strategy combined with the redundancy-aware operation can improve the contribution of the two by 2.5\%.
This proves that the differential learning and the redundancy-aware operation have mutually
reinforcing effects within the model. 
The combination of all proposed components 
brings the best performance, which significantly improves the performance by
4.0\% compared to the baseline model.
\section{Conclusion}
This paper proposes a lip reading model (RAL) that learns differential information from the symmetrical left and right half-lip views. 
We introduce a differential learning strategy with symmetric views to better capture the identical and different information of the left and right half-lips. Given that lip reading images contain a large amount of redundant information, we propose the redundancy-aware operation to reduce redundant information. To fully learn the relationship between symmetrical views, we design an adaptive cross-view interaction module. Experimental results fully validate the effectiveness of the proposed lip reading model.

\bibliographystyle{IEEEtran}
\bibliography{IEEEabrv,sample-base}

\end{document}